\documentclass[letterpaper]{article} 
\usepackage{aaai2027}
\usepackage[hyphens]{url}  
\usepackage{graphicx} 
\usepackage{natbib}  
\usepackage{caption} 
\usepackage{booktabs}
\usepackage{tabularx}
\usepackage{array}
\usepackage{verbatim}
\usepackage{makecell}
\usepackage{threeparttable}
\usepackage{longtable}
\usepackage{ragged2e}
\usepackage{multirow}
\usepackage{amsmath}
\usepackage{algorithm}
\usepackage{algorithmic}

\usepackage{newfloat}
\usepackage{listings}
\DeclareCaptionStyle{ruled}{labelfont=normalfont,labelsep=colon,strut=off} 
\floatstyle{ruled}
\newfloat{listing}{tb}{lst}{}
\floatname{listing}{Listing}

\usepackage{booktabs}

\title{The Uneven Impact of Generative AI on Student Learning: Examining the Roles of Reliance, Evaluation Literacy, and Course Policy in AI-related Courses}

\author{Lydia Manikonda\textsuperscript{\rm 1}, Mei Si, Sirajam Munira, 
Oshani Seneviratne, Kristin Bennett}
\affiliations{%
\textsuperscript{\rm 1}Corresponding author\\
Rensselaer Polytechnic Institute (RPI) \\ Troy, New York, USA\\
\{manikl, sim, munirs, senevo, bennek\}@rpi.edu}

\begin{document}

\maketitle

\begin{abstract}
Generative artificial intelligence (GenAI) is changing how students learn, yet the roles of course context, cognitive reliance, evaluation literacy, and early reliance remain underexplored. Using survey responses from 118 students across 12 AI-related courses at our institution, we examined differences in GenAI use and perceived learning experiences. We identified four user clusters: high-use students reporting many benefits, light users reporting less reliance and fewer benefits, and two moderate-use groups reporting different levels of benefit. We also found significant differences between free- and premium-version users, single- and multiple-tool users, and students experiencing different instructor policies. In multivariable regression models, academic benefit was associated with early reliance and academic task support; positive impact was associated with cognitive reliance, academic task support, confidence in GenAI reliability, and instructor policy; and negative impact was associated with early reliance and attitudinal change. The association between early reliance and negative impact became stronger as evaluation literacy increased. Finally, perceptions of GenAI-enhanced learning appear to reflect cognitive, performance, and self-efficacy benefits, while concerns about stress and diminished critical thinking are associated with lower perceived learning benefits. These findings suggest that institutions need better policies to address such inequities so that institutions can enable students to benefit from increasingly capable AI systems.




\end{abstract}


\section{Introduction} 

GenAI has rapidly become part of everyday learning at higher education institutions. Students use large language models (LLMs) to explain unfamiliar concepts, debug programs, summarize readings, brainstorm ideas, and complete academic work~\cite{cena2026studying}. In computing and AI-related courses, this development creates an especially consequential educational tension due to the ever-increasing performance quality of GenAI in such course-related topics. The same systems that students study as technical objects are increasingly used as partners in the process of studying. Existing literature has consistently found enthusiasm about GenAI's usefulness alongside concerns about inaccuracy, academic integrity, overreliance, and reduced engagement with learning~\cite{kim2025examining,marshik2025student,sah2025generative,destefano2026understanding}. Access is uneven as well: some students use free versions of a single system, while others can draw on premium models or several tools with different capabilities. These differences may shape not only how frequently students use GenAI, but also which learning activities it can support and how much benefit students perceive. 

Despite this growing literature, three gaps remain. First, much of the evidence comes from broad university populations or from interventions structured around a particular tool or task. 
Second, reliance is often treated as a single continuum. This can obscure an important distinction between using GenAI during cognitively substantive activities and consulting it before attempting a problem independently or seeking conventional forms of help. Third, evidence about students' confidence, access conditions, and perceived learning outcomes is rarely considered within a common empirical framework. We address these gaps through a survey of undergraduate and graduate students enrolled in 12 AI- and data-intensive courses at Rensselaer Polytechnic Institute (RPI), an R1 university in the United States, spanning artificial intelligence, machine learning, data analytics, decentralized systems, AI ethics, biostatistics, and economic analysis. 

The study measures academic task support, cognitive and early reliance, evaluation literacy, attitudinal change, confidence without GenAI, GenAI reliability, GenAI policy clarity, positive and negative impacts, academic benefit, limited learning impact, and course-supported effective GenAI use. We define \emph{cognitive reliance} as using GenAI to recall, organize, combine, evaluate, and apply information or solve problems. We distinguish this from \emph{early reliance}: consulting GenAI before independent thought, traditional search, or instructor support. We examine how these behaviors relate to students' perceived learning and whether their experiences differ according to access, number of tools used, academic level, and course context. 
Accordingly, we ask:
\begin{enumerate}
    \item \textbf{RQ1:} What distinct patterns of GenAI use, cognitive reliance, and perceived academic benefits and consequences emerge among students in AI-related courses?
    \item \textbf{RQ2:} How do factors, including access to free vs. premium GenAI tools, use of single vs. multiple tools, and instructor policies, relate to students' GenAI learning experiences?
    \item \textbf{RQ3:} What factors are associated with students' perceptions of overall improved learning through GenAI?
\end{enumerate}

\section{Related Work}

Prior research shows that GenAI can support learning while also raising concerns about cognitive offloading, overconfidence, and reduced independent engagement. Students' experiences may further vary with confidence, evaluation practices, and instructional context, but these dimensions are often studied separately or in broad university populations. Among the studies reviewed, we did not identify a survey-based analysis that jointly examined GenAI use, learning support, cognitive reliance, evaluation practices, and course policy among students in AI-related courses.

\subsubsection{Student Use and Perceptions of GenAI}
University students use GenAI for academic activities such as seeking explanations, finding information, brainstorming, summarizing, revising, programming, and solving problems. Students associate these uses with greater efficiency, access to assistance, confidence, and work quality, while also expressing concerns about inaccurate information, academic integrity, dependence, and reduced engagement~\cite{kim2025examining,marshik2025student,sah2025generative,destefano2026understanding}. These findings indicate that GenAI use is neither uniformly beneficial nor harmful. However, most surveys examine broad university populations, providing limited evidence about how GenAI use, perceived support, and learning outcomes relate within AI-related courses.

\subsubsection{Confidence, Cognitive Reliance, and Critical Evaluation}
Greater confidence in GenAI has been associated with less self-reported critical-thinking effort, whereas greater confidence in one's own abilities has been associated with more critical engagement~\cite{lee2025impact}. Critical thinking in AI-assisted work increasingly involves verifying, comparing, and integrating generated outputs, suggesting that self-confidence, confidence in GenAI, and critical evaluation are related but distinct. GenAI can reduce the cognitive demands of academic work, but excessive reliance may contribute to cognitive offloading and an illusory sense of mastery~\cite{wang2026generative}. The timing and structure of assistance also matter: unrestricted assistance may improve performance while GenAI is available without producing corresponding gains in independent performance, whereas learning-oriented safeguards can reduce this effect~\cite{bastani2025generative}. Together, this literature suggests that reliance should not be treated as a single construct. Using GenAI to organize information, decompose problems, evaluate work, or combine ideas may represent a different form of reliance from consulting GenAI before independent thought, traditional search, or instructor support. Existing surveys rarely distinguish these patterns or examine how they relate to confidence and evaluation practices.

\subsubsection{Instructor Policy and Student Practices}
Course policies provide important context for students' GenAI practices. Students and faculty often share general concerns about GenAI but differ in their perceptions of its use, acceptability, and effects on learning~\cite{kim2025examining,marshik2025student,sah2025generative,destefano2026understanding,manikonda2026policy}. Instructors may also misestimate students' GenAI use, trust, and preferences for AI assistance~\cite{saeli2026parallel}, while students frequently report uncertainty about institutional rules and acceptable uses~\cite{morari2026academic}. Course policy should therefore be considered both as a formal instructional condition and as something students perceive and interpret.

\section{Methodology}

\subsection{Study Design and Data Collection}

We surveyed students enrolled in AI- and data-analytics-related courses at our institution in the United States during Spring 2026. Students were recruited
through course instructors and in-class announcements. Participation was
voluntary and anonymous, with no monetary compensation, and the study received
approval from the institution's Institutional Review Board. A total of 143 complete responses were received, and the final sample included 118 undergraduate and graduate students from 12 AI-related courses across multiple disciplines.

The survey contained 60 Likert-type items covering GenAI use, perceived
impacts, cognitive reliance, evaluation practices, academic benefits,
attitudes, and course-policy experiences. These items were organized into
eight multi-item constructs and five single-item measures. Likert-type responses were coded on a 1-4 scale, with higher values indicating higher levels of the measured response (e.g., agreement, frequency, confidence, or importance); ``Not Applicable'' responses were coded as the scale midpoint of 2.5. Multi-item scores
were calculated by averaging their included items.

Internal consistency was assessed using Cronbach's alpha, with higher values
indicating that the items within a construct produced more consistent
responses. Table~\ref{tab:construct_measures} summarizes the measures, their
roles in the regression models, reliability, and descriptive statistics.
Additional demographic, access, tool-use, and course-policy variables were
used for the group comparisons described in
Section~\ref{sec:item-level}. The complete survey instrument is provided in Supplementary Table S1.

\begin{table*}[t]
\centering
\caption{Survey measures, internal consistency, and descriptive statistics.}
\label{tab:construct_measures}

\begin{threeparttable}
\scriptsize
\setlength{\tabcolsep}{3pt}
\renewcommand{\arraystretch}{1.10}

\begin{tabularx}{\textwidth}{
    @{}
    >{\raggedright\arraybackslash}p{2.65cm}
    >{\centering\arraybackslash}p{0.75cm}
    X
    >{\centering\arraybackslash}p{1.10cm}
    >{\centering\arraybackslash}p{0.70cm}
    >{\centering\arraybackslash}p{0.70cm}
    @{}
}
\toprule
\textbf{Measure} &
\textbf{Items} &
\textbf{Included content} &
\textbf{Cronbach's \(\boldsymbol{\alpha}\)} &
\textbf{Mean} &
\textbf{SD} \\
\midrule

\multicolumn{6}{l}{\textbf{Dependent variables}} \\[1mm]

Academic benefit
& 6
& Learning confidence; workload management; new perspectives; work quality;
grades or performance; retention of course material
& .876 & 2.48 & .78 \\

Positive impact
& 9
& Time savings; idea generation; progress; assistance when stuck; work quality;
creativity; beyond-expertise tasks; skill confidence; overall learning
& .892 & 2.90 & .59 \\

Negative impact
& 7
& Lower learning confidence; stress; unfair advantage; reduced motivation;
harmed critical thinking; fear of GenAI use; difficulty adapting output
& .768 & 2.03 & .55 \\

Learned GenAI
& 1
& Course helped the student to use GenAI more effectively
& -- & 2.70 & .80 \\

\midrule
\multicolumn{6}{l}{\textbf{Independent variables}} \\[1mm]

Academic task support
& 12
& Learning; information seeking; summarizing; brainstorming; revising;
completing and checking assignments; further thinking; course clarification;
practice activities; direct content generation
& .911 & 2.97 & .55 \\

Cognitive reliance
& 6
& Recalling, organizing, combining, and evaluating information; solving and
decomposing problems
& .864 & 2.76 & .66 \\

Evaluation literacy
& 8
& Evaluation confidence and reflection; checking sources, concepts,
organization, logic, synthesis, and task requirements
& .840 & 3.52 & .49 \\

Early reliance
& 4
& Consulting GenAI before traditional search, independent thinking, or
instructor support; preferring GenAI explanations
& .722 & 2.22 & .72 \\

Attitudinal change
& 3
& Changes in attitudes toward GenAI and interactions with instructors or peers
& .683 & 2.24 & .70 \\

Confidence without GenAI
& 1
& Confidence in completing tasks without GenAI
& -- & 3.28 & .63 \\

GenAI reliability
& 1
& Confidence in the reliability of GenAI output
& -- & 2.56 & .83 \\

GenAI policy clarity
& 1
& Clarity of guidelines for appropriate and inappropriate GenAI use
& -- & 2.92 & 1.01 \\

Limited GenAI impact
& 1
& Perception that GenAI had little impact on overall learning
& -- & 2.04 & .84 \\

\bottomrule
\end{tabularx}

\begin{tablenotes}[flushleft]
\scriptsize
\item \textit{Note.} \(N=118\) for all measures. Scores used a 1--4 response
scale, with Not Applicable coded as 2.5. Item wording is abbreviated for space.
Cronbach's \(\alpha\) is reported only for multi-item constructs.
\end{tablenotes}

\end{threeparttable}
\end{table*}

\begin{figure}[t]
    \centering
    \includegraphics[width=\linewidth]{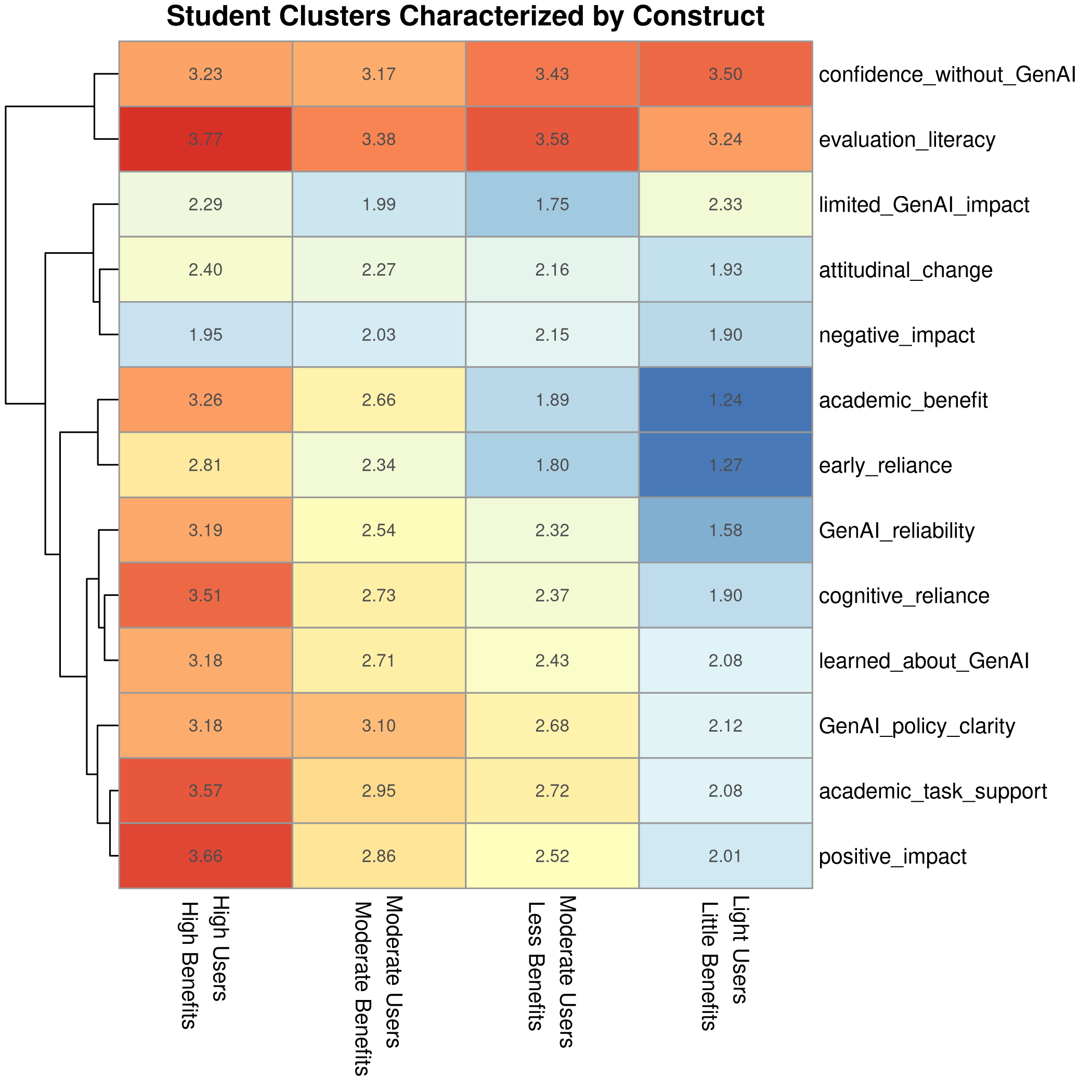}
    \caption{Mean construct scores across the four student clusters.}
    \label{fig:heatmap}
\end{figure}

\subsection{Analytic Approach}

We used four complementary analyses. Clustering identified patterns of GenAI
use and experience; group comparisons examined differences by student, access,
and course characteristics; multiple regression estimated adjusted
associations with the four learning outcomes; and an ensemble model explored
potential nonlinear relationships with overall perceived learning.

\subsubsection{Clustering}

To identify patterns of GenAI use and experience, we applied hierarchical
agglomerative clustering with Ward's method to students' responses to the 60
Likert-type items. The four-cluster solution was selected by inspecting the
resulting dendrogram and heatmap and considering the interpretability of the
response profiles. Cluster-level construct means were then used to describe
the four profiles. As described above, Not Applicable responses were coded as
2.5 before clustering.

\subsubsection{Item-Level Student Group Comparisons}
\label{sec:item-level}

We compared individual Likert-type responses across groups defined by academic
level, school, course activity, GenAI access, number of tools used, and
perceived instructor policy. The comparisons included graduate versus
undergraduate students; students from four schools; lecture-only courses versus
courses with an additional GenAI-related activity; premium-access versus
free-only users; self-paid premium access versus premium access provided by
another source; and single-tool versus multi-tool users.

Instructor policy was analyzed across all five response categories and,
separately, across the three substantive policy categories: encouraged,
allowed with limits, and discouraged. Two-group comparisons used two-sided
Mann--Whitney \(U\) tests with rank-biserial correlation as the effect size. Comparisons involving more than two groups used Kruskal--Wallis tests, followed
when appropriate by Dunn post-hoc comparisons with Holm adjustment.
Benjamini--Hochberg false-discovery-rate correction was applied within each
group-comparison family, with adjusted \(p<.05\) considered statistically
significant.

\subsubsection{Multiple Regression}
\label{sec:regressionmethod}

Separate multiple linear regressions were estimated for academic benefit,
positive impact, negative impact, and course-supported GenAI learning. All
predictors were entered simultaneously, with ``allowed with limits'' serving
as the reference category for instructor policy. All variables were
standardized, and each model included all 118 respondents. HC3 standard errors were used for the primary two-sided \(p\)-values, with
course-clustered standard errors calculated as a robustness check. Raw
\(p\)-values are reported without multiple-testing correction. Four
theoretically motivated interactions were also tested in separate extensions
of the corresponding regression models.

\subsubsection{Ensemble Learning and Feature Rankings}

As a complementary exploratory analysis, we trained a gradient-boosting
classifier to examine potential nonlinear relationships between the survey
measures and students' responses regarding whether GenAI improved their
learning. We then used Shapley (SHAP) values~\cite{lundberg2017unified} to rank the features according to their contributions to the model and to show whether
higher feature values were associated with higher or lower model predictions.
This analysis was used to explore feature importance rather than to replace
the adjusted associations estimated by the regression models.

\section{Results}

\subsection{Clustering }

Clustering revealed four distinct types of students. Figure~\ref{fig:heatmap} summarizes each of the clusters by their average Likert values for the 13 different constructs across the 118  subjects.  The rows correspond to the measures defined in  Table~\ref{tab:construct_measures}. The four clusters describe four different groups of students with high, medium, and low AI usage characterized by their ranking order on the constructs that describe their AI usage Academic Task Support, Cognitive Reliance, and Early Reliance.  Each group exhibits High, Medium, Less, and Little Benefits as indicated by  positive impact and academic benefits. Thus, we refer to the clusters as High Usage/High Benefits, Moderate Usage/Moderate Benefits, Moderate Usage/Less Benefits, and Less Usage/Less Benefits. We see that these constructs correlate with the course specific questions on GenAI clarity and whether they learned GenAI in the course. This suggests that students in courses that teach GenAI with clear usage policies tend to use GenAI for more academic tasks, experience more positive impacts, and are more confident about using GenAI reliably. The greater the usages and benefits of AI reported by students, the more students report that they experienced that AI changed in their interactions with peers and faculty.  

The clustering shows that confidence in completing tasks without GenAI is inversely correlated with Usage and Benefits, suggesting that higher AI usage and benefits creates students in AI classes that are more reliant on GenAI as part of learning.  All students report high Evaluation Literacy, but rather interestingly the students with Moderate Usage/Less Benefits show a more enhanced grasp of AI evaluation that Medium Usage-Medium Benefits cluster.  The two medium usage groups illustrate some interesting correlations.  They are more likely to say that GenAI impacted their learning as indicated by their disagreement with the statement ``GenAI has no impact on learning''. It appears that moderate usage users also report slightly more negative impacts, although all users report relative few negative impacts on average.  

The clustering results suggest that students' AI usage, benefits, and impacts  vary unevenly across different AI related courses and that these may be related to  course specific contexts such as whether GenAI is taught and GenAI policies are clear.  Note that usage and benefits may also be partially explained by the percentage of graduate students within each cluster:  High Usage/High Benefits (46.4\%),  Moderate Users/Moderate Benefits (32.6\%), Moderate Users/Less Benefits (24.1\%), and Low Usage/Low Benefits (16.6\%). 

\subsection{Item-Level Differences Across Student Groups}
Table~\ref{tab:individual_significance} summarizes individual survey items that are significant after FDR correction. No FDR-significant item-level differences were observed by academic level, school, course activity, or whether students paid for GenAI access themselves versus receiving access from another source. In contrast, significant differences emerged for premium versus free-only access, single- versus multi-tool use, and perceived instructor policy. The clearest pattern emerged for single- versus multi-tool use. Multi-tool users reported higher responses for nearly all significant items, spanning academic-task support, positive impact, cognitive reliance, early reliance, and academic benefit. The largest differences were for improved academic-work quality ($r_{rb}=-0.53$), retention of course material ($-0.49$), using GenAI to evaluate work ($-0.48$), improving work quality ($-0.45$), and breaking down complex problems ($-0.44$). The one exception was confidence without GenAI: single-tool users reported greater confidence in completing tasks without GenAI ($r_{rb}=+0.25$). This contrast suggests that broader tool use was associated with greater perceived support and reliance on GenAI, but not with greater confidence in working independently without it.

\begin{table*}[!t]
\centering
\caption{Significant Individual-Item Differences Across GenAI Access, Tool Use, and Instructor Policy}
\label{tab:individual_significance}

\begin{threeparttable}

\scriptsize
\setlength{\tabcolsep}{2.0pt}
\renewcommand{\arraystretch}{0.82}

\begin{tabularx}{\textwidth}{
    @{}
    >{\raggedright\arraybackslash}p{2.65cm}
    >{\centering\arraybackslash}p{0.85cm}
    X
    >{\centering\arraybackslash}p{2.35cm}
    >{\centering\arraybackslash}p{2.35cm}
    >{\centering\arraybackslash}p{2.35cm}
    @{}
}
\toprule
\textbf{Construct} &
\textbf{Item} &
\textbf{Survey item} &
\makecell{\textbf{Premium access}\\\textbf{vs. free-only}} &
\makecell{\textbf{Single-tool}\\\textbf{vs. multi-tool}} &
\makecell{\textbf{Perceived}\\\textbf{instructor policy}} \\
\midrule

Academic task support
& Q14\_1 & Learning about new topics
& -- & .0016\textsuperscript{**} (-.38) & -- \\

& Q14\_2 & Finding specific facts or information
& .0354\textsuperscript{*} (+.27) & .0326\textsuperscript{*} (-.24) & -- \\

& Q14\_3 & Discovering new insights from information or data
& -- & .0096\textsuperscript{**} (-.30) & -- \\

& Q14\_4 & Generating summaries of longer content
& -- & .0034\textsuperscript{**} (-.35) & .0494\textsuperscript{*} (.07) \\

& Q14\_5 & Brainstorming or organizing ideas before starting an assignment
& .0354\textsuperscript{*} (+.28) & .0006\textsuperscript{***} (-.41) & .0157\textsuperscript{*} (.11) \\

& Q14\_6 & Improving or rewriting assignments
& -- & .0184\textsuperscript{*} (-.28) & -- \\

& Q14\_7 & Completing assignments or significant portions of them
& -- & .0112\textsuperscript{*} (-.30) & .0458\textsuperscript{*} (.07) \\

& Q14\_8 & Checking whether work meets rules, constraints, or formatting requirements
& -- & .0120\textsuperscript{*} (-.29) & -- \\

& Q14\_10 & Summarizing lecture slides or notes to clarify materials
& -- & .0016\textsuperscript{**} (-.40) & .0458\textsuperscript{*} (.06) \\

& Q14\_11 & Generating practice tests or quizzes
& -- & -- & .0494\textsuperscript{*} (.07) \\

& Q14\_12 & Generating content for direct use in assignments
& -- & -- & .0157\textsuperscript{*} (.10) \\

\midrule

Positive impact
& Q18\_1 & Save time
& .0151\textsuperscript{*} (+.31) & .0018\textsuperscript{**} (-.36) & -- \\

& Q18\_2 & Generate ideas
& -- & .0027\textsuperscript{**} (-.37) & -- \\

& Q18\_3 & Make progress in work
& -- & -- & .0157\textsuperscript{*} (.09) \\

& Q18\_4 & Continue work when stuck
& .0043\textsuperscript{**} (+.38) & .0023\textsuperscript{**} (-.36) & -- \\

& Q18\_5 & Improve quality of work
& .0043\textsuperscript{**} (+.39) & .0003\textsuperscript{***} (-.45) & .0367\textsuperscript{*} (.08) \\

& Q18\_6 & Be more creative
& -- & .0042\textsuperscript{**} (-.35) & -- \\

& Q18\_7 & Complete tasks beyond current expertise
& .0012\textsuperscript{**} (+.46) & .0042\textsuperscript{**} (-.35) & -- \\

& Q23\_1 & More confident in skillset after using GenAI
& -- & -- & .0017\textsuperscript{**} (.19) \\

& Q25\_4 & GenAI improved learning
& -- & .0079\textsuperscript{**} (-.30) & -- \\

\midrule

Confidence without AI
& Q15 & Confident completing tasks without GenAI
& -- & .0184\textsuperscript{*} (+.25) & -- \\

\midrule

GenAI reliability
& Q16 & Confident in GenAI reliability
& .0117\textsuperscript{*} (+.34) & .0034\textsuperscript{**} (-.33) & .0157\textsuperscript{*} (.10) \\

\midrule

Cognitive reliance
& Q19\_2 & Organize or summarize ideas
& -- & .0016\textsuperscript{**} (-.39) & -- \\

& Q19\_3 & Combine ideas to create new meanings
& .0354\textsuperscript{*} (+.30) & .0061\textsuperscript{**} (-.34) & .0157\textsuperscript{*} (.10) \\

& Q19\_4 & Evaluate and quality-check work
& .0120\textsuperscript{*} (+.34) & .0001\textsuperscript{***} (-.48) & -- \\

& Q19\_5 & Solve new problems
& -- & .0016\textsuperscript{**} (-.38) & .0367\textsuperscript{*} (.08) \\

& Q19\_6 & Break down complex problems
& .0349\textsuperscript{*} (+.29) & .0003\textsuperscript{***} (-.44) & .0458\textsuperscript{*} (.07) \\

\midrule

Learned GenAI
& Q23\_6 & Course helped use GenAI more effectively
& .0349\textsuperscript{*} (+.30) & -- & .0113\textsuperscript{*} (.14) \\

\midrule

Early reliance
& Q24\_1 & Consult GenAI before traditional search
& .0323\textsuperscript{*} (+.31) & .0036\textsuperscript{**} (-.34) & -- \\

& Q24\_9 & GenAI explains concepts better than instructor
& -- & .0080\textsuperscript{**} (-.31) & -- \\

& Q24\_14 & Use GenAI before reaching instructor
& -- & .0078\textsuperscript{**} (-.31) & -- \\

\midrule

Academic benefit
& Q24\_2 & Use GenAI to boost confidence
& -- & .0090\textsuperscript{**} (-.30) & -- \\

& Q24\_3 & GenAI helps manage time/workload
& -- & .0120\textsuperscript{*} (-.29) & -- \\

& Q24\_4 & GenAI exposes new ideas/perspectives
& -- & .0016\textsuperscript{**} (-.39) & -- \\

& Q24\_5 & GenAI improves academic work quality
& .0043\textsuperscript{**} (+.40) & $<$.0001\textsuperscript{***} (-.53) & -- \\

& Q24\_7 & GenAI improves grades/performance
& -- & .0060\textsuperscript{**} (-.33) & .0358\textsuperscript{*} (.08) \\

& Q24\_8 & GenAI helps retain course materials
& -- & .0001\textsuperscript{***} (-.49) & .0367\textsuperscript{*} (.08) \\

\bottomrule
\end{tabularx}

\begin{tablenotes}[flushleft]
\scriptsize
\item \textit{Note.}
Only individual items with at least one FDR-significant comparison across the three variables shown are included.
Each reported cell gives the Benjamini--Hochberg FDR-adjusted
$p$-value followed by the effect size in parentheses.
\textsuperscript{*}$p_{\mathrm{FDR}}<.05$,
\textsuperscript{**}$p_{\mathrm{FDR}}<.01$,
and \textsuperscript{***}$p_{\mathrm{FDR}}<.001$.
A dash indicates no significant difference after FDR correction.

For the two-group comparisons, effect size is the rank-biserial
correlation ($r_{rb}$).
For \textit{Premium access vs. free-only access},
a positive value indicates higher responses among students with premium
access, whereas a negative value indicates higher responses among
free-only users.
For \textit{Single-tool vs. multi-tool},
a positive value indicates higher responses among single-tool users,
whereas a negative value indicates higher responses among multi-tool users.
For \textit{Perceived instructor policy}, eligible responses were
\textit{Encouraged}, \textit{Allowed with limits}, and \textit{Discouraged};
effect size is epsilon-squared ($\epsilon^2$), which indicates magnitude
but not direction.
\end{tablenotes}

\end{threeparttable}
\end{table*}

Premium access showed fewer but generally positive differences. Premium users reported greater use of GenAI for finding information and brainstorming, and higher perceived benefits for saving time, continuing work when stuck, improving work quality, completing tasks beyond their expertise, and improving academic-work quality. The largest premium-access difference was for completing tasks beyond current expertise ($r_{rb}=0.46$), followed by improved academic-work quality ($0.40$). The three-category substantive policy comparison produced the FDR-significant differences; the five-category analysis yielded no additional substantive pattern. Perceived instructor policy was associated with several items, although effect sizes were generally small and ranged from  $\epsilon^2=0.06$ to $0.19$. Differences appeared in task use, confidence in GenAI reliability, cognitive reliance, perceived learning and academic benefits, and whether the course helped students use GenAI more effectively. The largest policy-related effect was for confidence in one's skillset after using GenAI ($\epsilon^2=0.19$). Because the omnibus policy test does not indicate direction, these results reflect variation across policy groups rather than a specific ordering among encouraged, limited, and discouraged use.

\subsection{Regression Analysis}
\label{sec:regressionmethod}


\begin{table*}[t]
\centering
\caption{Significant main effects from the multiple regression models.}
\label{tab:regression-associations}
\small
\begin{tabular}{llrrr}
\toprule
Outcome & Predictor & \(\beta\) & HC3 SE & \(p\) \\
\midrule
Academic benefit
& Early reliance & .301 & .084 & \(<.001^{**}\) \\
& Academic task support & .294 & .108 & \(.008^{**}\) \\
\addlinespace
Positive impact
& Cognitive reliance & .495 & .130 & \(<.001^{**}\) \\
& Academic task support & .242 & .099 & \(.016^{*}\) \\
& Instructor policy: discouraged & \(-.138\) & .057 & \(.017^{*}\) \\
& Confidence in GenAI reliability & .196 & .083 & \(.020^{*}\) \\
\addlinespace
Negative impact
& Attitudinal change & .386 & .108 & \(<.001^{**}\) \\
& Early reliance & .402 & .138 & \(.004^{**}\) \\
\addlinespace
Course-supported GenAI learning
& Instructor policy: discouraged & \(-.186\) & .059 & \(.002^{**}\) \\
& Attitudinal change & .242 & .116 & \(.039^{*}\) \\
\bottomrule
\end{tabular}

\smallskip
\begin{minipage}{0.92\textwidth}
\footnotesize
\textit{Note.} Coefficients are standardized and mutually adjusted.
\(^{*}p<.05\); \(^{**}p<.01\). Reported \(p\)-values are raw, two-sided HC3
values.
\end{minipage}
\end{table*}

\subsubsection{Main Effects}

All four regression models included the full sample of 118 respondents. The
predictor set explained 68.4\% of the variance in academic benefit
(adjusted \(R^2=.641\)), 79.7\% of the variance in positive impact
(adjusted \(R^2=.771\)), 30.0\% of the variance in negative impact
(adjusted \(R^2=.212\)), and 36.9\% of the variance in course-supported GenAI
learning (adjusted \(R^2=.290\)). The models therefore accounted for more
variation in academic benefit and positive impact than in negative impact or
course-supported GenAI learning. Table~\ref{tab:regression-associations} presents the significant main effects
from the mutually adjusted regression models. Each coefficient represents the
association between one predictor and the outcome while holding the other
predictors constant. Only coefficients with raw HC3 \(p<.05\) are shown; no
multiple-testing correction was applied. The significant main effects were as follows:

\begin{itemize}
    \item \textbf{Academic benefit.} Self-reported academic benefit was
    positively associated with early reliance
    (\(\beta=.301\), \(p<.001\)) and academic task support
    (\(\beta=.294\), \(p=.008\)).

    \item \textbf{Positive impact.} Self-reported positive impact was
    positively associated with cognitive reliance
    (\(\beta=.495\), \(p<.001\)), academic task support
    (\(\beta=.242\), \(p=.016\)), and confidence in GenAI reliability
    (\(\beta=.196\), \(p=.020\)). Positive impact was lower when instructor
    policy discouraged GenAI rather than allowed it with limits
    (\(\beta=-.138\), \(p=.017\)).

    \item \textbf{Negative impact.} Self-reported negative impact was
    positively associated with early reliance
    (\(\beta=.402\), \(p=.004\)) and attitudinal change
    (\(\beta=.386\), \(p<.001\)).

    \item \textbf{Course-supported GenAI learning.} Self-reported
    course-supported GenAI learning was positively associated with attitudinal
    change (\(\beta=.242\), \(p=.039\)). It was lower when instructor policy
    discouraged GenAI rather than allowed it with limits
    (\(\beta=-.186\), \(p=.002\)).
\end{itemize}

Course-clustered inference supported most of these findings. The association
between academic task support and positive impact was no longer significant
after standard errors were clustered by course (\(p=.060\)); all other main
effects reported in Table~\ref{tab:regression-associations} remained
significant at \(p<.05\).

\subsubsection{Interaction Effects}

\begin{figure}[t]
    \centering
    \includegraphics[width=\linewidth]
    {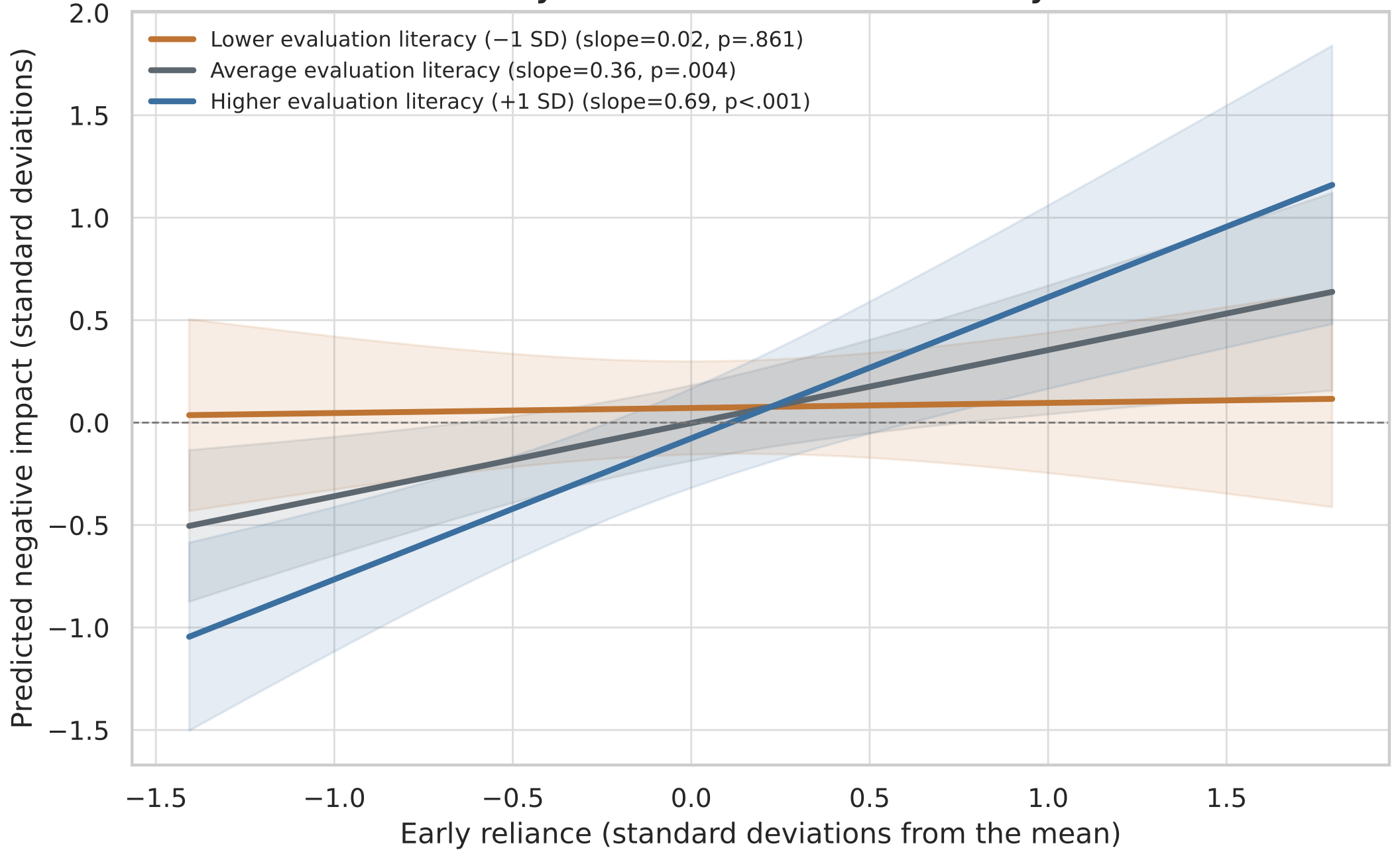}
    \caption{Association between early reliance and negative impact at lower,
    average, and higher levels of evaluation literacy. Shaded regions show
    95\% HC3 confidence intervals.}
    \label{fig:early-reliance-evaluation-interaction}
\end{figure}

Four interactions were tested to determine whether the association between a
predictor and an outcome changed according to the level of another predictor:
academic task support \(\times\) evaluation literacy, early reliance
\(\times\) evaluation literacy, early reliance \(\times\) policy clarity, and
cognitive reliance \(\times\) confidence without GenAI. The interaction between early reliance and evaluation literacy was significant
for negative impact (\(\beta=.307\), \(p<.001\)). As shown in
Figure~\ref{fig:early-reliance-evaluation-interaction}, early reliance was not
associated with negative impact at lower evaluation literacy
(\(-1\) SD; \(b=.025\), \(p=.861\)). The association was positive at average
evaluation literacy (\(b=.357\), \(p=.004\)) and stronger at higher evaluation
literacy (\(+1\) SD; \(b=.688\), \(p<.001\)). Thus, early reliance was
associated with greater negative impact primarily among students with average
or higher evaluation literacy. This interaction remained significant when standard errors were clustered by
course (\(p<.001\)). The other three interactions were not significant in the
primary HC3 analysis. The early reliance \(\times\) policy clarity interaction
was significant only under course-clustered inference (\(p=.026\)) and was
therefore not treated as a primary finding.

\subsection{Impact of Factors on Learning with GenAI}
\begin{figure}[h]
    \centering
    \includegraphics[width=\linewidth]{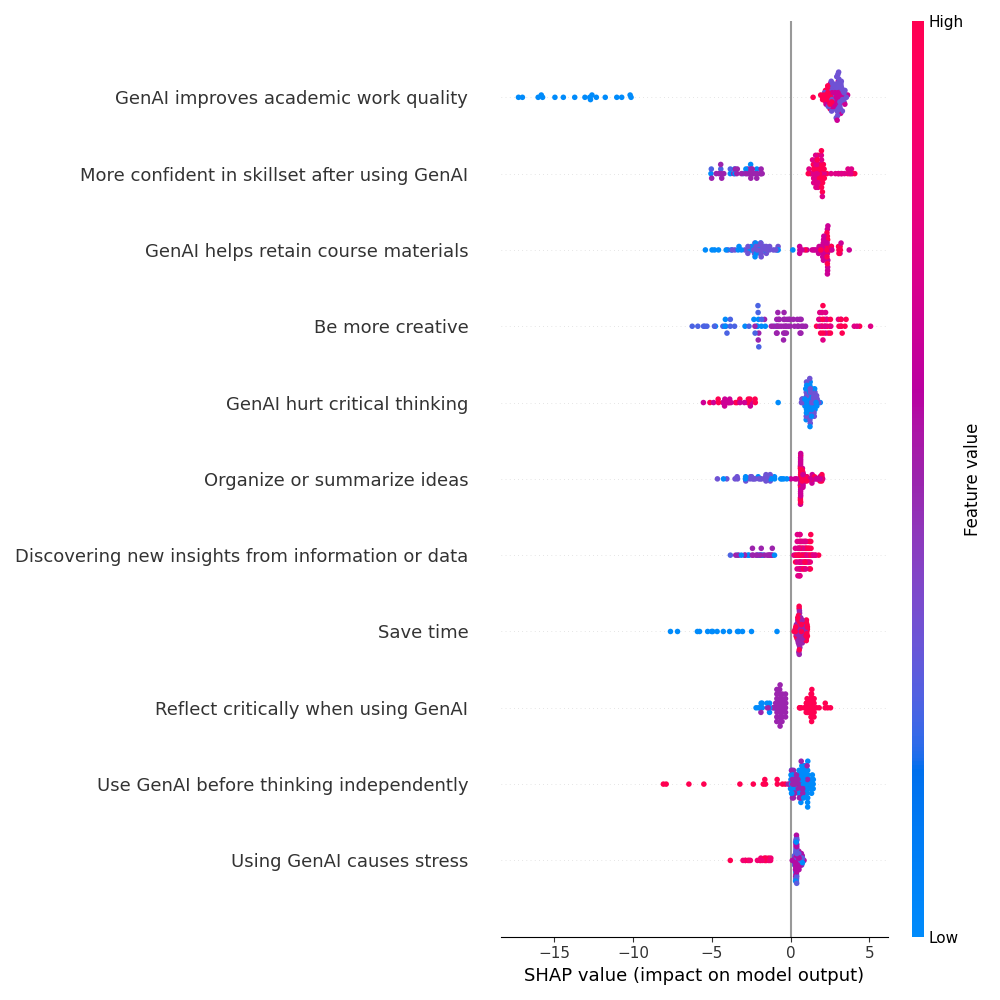}
    \caption{SHAP summary plot illustrating the top features and their direction of influence on the outcome variable ``Overall, GenAI improved learning''}
    \label{fig:Q44-4}
\end{figure}

To further understand factors that are highly correlated with the overall influence of GenAI impacting learning, 
We utilized all the constructs except the target item present in our survey results and train an ensemble learning algorithm via Gradient Boosting Classifier to predict the outcome variable -- if ``overall, GenAI improved learning.''  The SHAP summary plot in Figure~\ref{fig:Q44-4} shows the feature rankings and their direction of influence on overall how GenAI improved learning. Features are ranked vertically by their cumulative impact on the model. Each dot represents a single observation where, the color represents the feature value (red for high, blue for low), and the horizontal axis measures the SHAP value (impact on the prediction). For example, high values of strong belief that GenAI improved academic work quality (red dots) strongly correlates with a positive model output. 

The findings suggest that improved learning through GenAI is associated not simply with its use but with the role it plays in the learning process. SHAP results show that students report greater learning benefits when they perceive GenAI as augmenting their existing capabilities. This includes improving the quality of academic work, being more creative, helping organize and summarize ideas, increasing confidence in their skills, and facilitating the discovery of new insights. In contrast, perceived learning benefits are lower when students experience GenAI as interfering with their cognitive engagement, particularly when they believe it hurts their critical thinking abilities or causes stress. Other factors, such as using GenAI saves time, are also positively associated with positive experiences of using GenAI in learning. Collectively, these findings suggest that the educational value of GenAI seems to be dependent more on whether it supports and extends students' cognitive processes rather than substitutes for them. 

\section{Discussion}
This study examined how GenAI reliance, evaluation literacy, access conditions, and course policy relate to students' self-reported learning experience in AI-related coureses. The results show that the reliance on GenAI is not simply a linear dependence such as ``high'' or ``low'' but how they engage with it and evaluate its outputs. For instance, early reliance and attitudinal change were associated with greater negative impact. At the same time, premium-access and multiple-tool users reported greater benefits and engagement, but neither group reported significantly greater negative impact. These findings suggest that focusing explicitly on the frequency of usage alone might not be a reliable factor when looking into the impacts of GenAI. Through our insights, it is also uncovered that there are diverse types of tasks that are related to cognitive offloading that may not be harmful to critical thinking in general. For instance, organizing or summarizing ideas and discovering new insights are positively associated with overall improved learning with GenAI, but caution should be exercised when it comes to prolonged reliance on these tools. Also, for this type of offloading to be productive, students should have sufficient domain knowledge and evaluation literacy to determine what has been offloaded. 

One interesting aspect we found in our analysis was about confidence. Single GenAI-tool users reported greater confidence in completing tasks without GenAI than multi-tool users. This contrast suggests that broader tool use was associated with greater perceived support and reliance on GenAI but not with greater confidence in working independently without it. This should be a cautionary lesson for instructors and institutes in general on how to leverage GenAI in such a way that it does not harm the ability of independent thinking and confidence in solving tasks independently in environments and careers where there is not always a possibility of accessing GenAI tools. Similarly, self-reported academic benefits were associated with early reliance and academic task support, whereas positive impacts of GenAI use on academic experiences were attributed to cognitive reliance, academic task support, confidence in GenAI reliability, and less restrictive instructor policies. Negative impacts of GenAI use on academic experiences were attributed to early reliance and attitudinal change, with the association between early reliance and negative impacts becoming stronger as students' evaluation literacy increased. With these types of uneven impact of GenAI on students, We hope that these results will pave the way to eventually address such disparities. 

Even though the results are insightful our study has certain limitations. This study relies on survey responses from only one R1 university. The insights in this work reflect only of those students who chose to share their experiences and may not represent the broader student population from diverse types of academic institutions. The survey also focused on thirteen different analysis measures -- eight multi-item constructs and five single-item measures. Our insights are only limited to these thirteen measures and may not capture all the universally associated factors in the context of learning and relying on GenAI. However, we do believe that given the diverse set of students with diverse experiences of using GenAI, we strongly believe that these insights are robust and even though preliminary can be utilized to expand this survey to a larger set of course at multiple institutes which will be a part of our future work. 

\section{Conclusion}
This study demonstrates that the role of GenAI in student learning at higher education institutions cannot simply be characterized by whether and how students use these tools. Instead, learning experiences are uneven and are shaped by students' types of reliance, evaluation literacy, positive and negative learning impacts, and the contexts in which GenAI is used. The four user profiles further showed that students with similar levels of use and reliance did not necessarily report similar academic benefits or consequences. How students relied on GenAI also mattered. Students who used it to organize ideas, evaluate their work, or solve problems (\emph{cognitive reliance}) generally reported more positive academic experiences. In contrast, students who turned to GenAI before thinking independently, searching for information, or asking an instructor (\emph{early reliance}) reported both academic benefits and negative effects. 
As GenAI becomes increasingly embedded in higher education, we argue that educators should move beyond policies that simply permit or restrict its use. Policies should clarify appropriate uses, support equitable access, and help students use GenAI to extend rather than replace independent reasoning. The central question is not simply whether students should rely on GenAI, but when and how that reliance can support learning without weakening independent thinking and critical reasoning.

\section*{Acknowledgements}
This research was supported by a grant from Google.org via EMPIRE AI. We thank Dr. Josephine C. Seddon, PhD for her valuable feedback on our survey instrument. 

\footnotesize
\bibliography{references}

\clearpage
\onecolumn
\appendix 
\renewcommand{\thetable}{\thesection.\arabic{table}}
\setcounter{table}{0}

\section{Student Survey Instrument}

\scriptsize
\setlength{\tabcolsep}{3pt}
\renewcommand{\arraystretch}{0.92}

\setlength\LTleft{0pt}
\setlength\LTright{0pt}

\begin{longtable}{
    @{}
    >{\RaggedRight\arraybackslash}p{2.6cm}
    >{\RaggedRight\arraybackslash}p{1.5cm}
    >{\RaggedRight\arraybackslash}p{10.0cm}
    >{\RaggedRight\arraybackslash}p{1.8cm}
    @{}
}

\caption{Student Survey Instrument}
\label{tab:student_survey_instrument}\\

\toprule
\textbf{Section / Construct} &
\textbf{Item} &
\textbf{Question / Statement} &
\textbf{Response Format} \\
\midrule
\endfirsthead

\multicolumn{4}{c}
{{\tablename\ \thetable{} -- continued from previous page}}\\
\toprule
\textbf{Section / Construct} &
\textbf{Item} &
\textbf{Question / Statement} &
\textbf{Response Format} \\
\midrule
\endhead

\midrule
\multicolumn{4}{r}{Continued on next page}\\
\endfoot

\bottomrule
\endlastfoot

\textbf{Consent}
& Q1
& By selecting ``I agree'' below, I confirm that I am at least 18 years of age and that I consent to participation in the research study described above.
& Consent \\

\midrule

\textbf{Demographics}
& Q2
& Level of Study
& Categorical \\

& Q3
& School
& Categorical \\

& Q4
& Major
& Categorical \\

\midrule

\textbf{GenAI Knowledge}
& Q5
& How would you describe what Generative AI (GenAI) is?
& Open-ended \\

& Q6
& How would you describe how Generative AI (GenAI) can be used?
& Open-ended \\

\midrule

\textbf{Course Info}
& Q7
& Course name and code/number
& Open-ended \\

& Q8
& Semester and year
& Open-ended \\

& Q9
& Course features
& Multiple selection \\

& Q10
& Briefly describe the activities and assignments in this course (e.g., coding, writing, discussions, group projects, presentations, labs).
& Open-ended \\

\midrule

\textbf{GenAI Tools Used}
& Q11
& Which GenAI tools do you use in this course?
& Multiple selection \\

& Q11\_5\_TEXT
& Other GenAI tool (please specify).
& Open-ended \\

& Q12
& Which specific Generative AI tools have you used in this course?
& Multiple selection \\

& Q12\_5\_TEXT
& Other GenAI tool (please specify).
& Open-ended \\

\midrule

\textbf{Instructor Policy}
& Q13
& What is your understanding of the instructor's policy on GenAI use in this course?
& Categorical \\

\midrule

\textbf{Academic Task Support}
& Q14\_1
& GenAI is helpful in this course for learning about new topics.
& Likert \\

& Q14\_2
& GenAI is helpful in this course for finding specific facts or information.
& Likert \\

& Q14\_3
& GenAI is helpful in this course for discovering new insights from information or data.
& Likert \\

& Q14\_4
& GenAI is helpful in this course for generating summaries of longer content.
& Likert \\

& Q14\_5
& GenAI is helpful in this course for brainstorming or organizing ideas before starting an assignment.
& Likert \\

& Q14\_6
& GenAI is helpful in this course for improving or rewriting assignments.
& Likert \\

& Q14\_7
& GenAI is helpful in this course for completing assignments or significant portions of them.
& Likert \\

& Q14\_8
& GenAI is helpful in this course for checking whether work meets rules, constraints, or formatting requirements.
& Likert \\

& Q14\_9
& GenAI is helpful in this course for generating ideas to prompt thinking, even if not directly using any of the generated text.
& Likert \\

& Q14\_10
& GenAI is helpful in this course for summarizing lecture slides or notes to help explain or clarify course materials.
& Likert \\

& Q14\_11
& GenAI is helpful in this course for generating practice tests or quizzes to study for assessments in this course.
& Likert \\

& Q14\_12
& GenAI is helpful in this course for generating content (e.g., text, Python code, or images) for direct use in assignments with modifications.
& Likert \\

\midrule

\textbf{Confidence Without AI}
& Q15
& I feel confident in my ability to complete tasks without any use of GenAI.
& Likert \\

\midrule

\textbf{GenAI Reliability}
& Q16
& I feel confident in GenAI's reliability to complete the course tasks.
& Likert \\

\midrule

\textbf{Negative Impact}
& Q17
& I feel it challenging to adapt the content generated by GenAI to be used for this assignment in this course.
& Likert \\

\midrule

\textbf{Positive Impact}
& Q18\_1
& In this course, GenAI helps me to save time.
& Likert \\

& Q18\_2
& In this course, GenAI helps me to generate ideas.
& Likert \\

& Q18\_3
& In this course, GenAI helps me to make progress in my work.
& Likert \\

& Q18\_4
& In this course, GenAI helps me to continue in my work when I am stuck on next steps.
& Likert \\

& Q18\_5
& In this course, GenAI helps me to improve the quality of my work.
& Likert \\

& Q18\_6
& In this course, GenAI helps me to be more creative in my work.
& Likert \\

& Q18\_7
& In this course, GenAI helps me to complete tasks I do not yet have the expertise to complete on my own.
& Likert \\

\midrule

\textbf{Cognitive Reliance}
& Q19\_1
& In this course, I rely on GenAI when I need to recall facts and basic concepts.
& Likert \\

& Q19\_2
& In this course, I rely on GenAI when I need to organize or summarize ideas.
& Likert \\

& Q19\_3
& In this course, I rely on GenAI when I need to combine ideas to create new meanings.
& Likert \\

& Q19\_4
& In this course, I rely on GenAI when I need to evaluate and quality-check my work.
& Likert \\

& Q19\_5
& In this course, I rely on GenAI when I need to solve new problems.
& Likert \\

& Q19\_6
& In this course, I rely on GenAI when I need to break down complex problems.
& Likert \\

\midrule

\textbf{Evaluation Literacy}
& Q20
& I feel confident in my ability to evaluate the accuracy and quality of GenAI's outputs.
& Likert \\

& Q21
& I reflect on and think critically when using GenAI for my coursework.
& Likert \\

& Q22\_1
& How important do you think it is to cross-check GenAI outputs with other trusted sources?
& Likert \\

& Q22\_2
& How important do you think it is to evaluate whether the GenAI output is well-organized?
& Likert \\

& Q22\_3
& How important do you think it is to reflect on whether GenAI correctly applies key concepts included in the output?
& Likert \\

& Q22\_4
& How important do you think it is to evaluate whether the GenAI output ideas or arguments are logically coherent?
& Likert \\

& Q22\_5
& How important do you think it is to assess how well the GenAI output combines ideas to form new meanings?
& Likert \\

& Q22\_6
& How important do you think it is to determine whether the GenAI output meets my expectations?
& Likert \\

\midrule

\textbf{Positive Impact}
& Q23\_1
& I feel more confident in my skillset after using GenAI in this course.
& Likert \\

\textbf{Negative Impact}
& Q23\_2
& I feel less confident in my learning abilities after using GenAI in this course.
& Likert \\

& Q23\_3
& Using GenAI in this course causes me stress.
& Likert \\

& Q23\_4
& I worry that my peers use GenAI to gain an unfair advantage.
& Likert \\

\textbf{Attitudinal Change}
& Q23\_5
& My attitudes toward GenAI changed over time in this course.
& Likert \\

\textbf{Learned GenAI}
& Q23\_6
& This course helped me to use GenAI more effectively.
& Likert \\

\midrule

\textbf{Early Reliance}
& Q24\_1
& How frequently do you consult GenAI before using traditional search engines or other resources?
& Likert \\

\textbf{Academic Benefit}
& Q24\_2
& How frequently do you use GenAI to boost your confidence in learning?
& Likert \\

& Q24\_3
& How frequently do you find that GenAI helps you manage your time and workload effectively?
& Likert \\

& Q24\_4
& How frequently do you find that GenAI exposes you to ideas or perspectives you would not have otherwise considered?
& Likert \\

& Q24\_5
& How frequently do you find that GenAI improves the quality of your academic work?
& Likert \\

\textbf{Negative Impact}
& Q24\_6
& How frequently do you find that GenAI reduces your motivation to learn course materials?
& Likert \\

\textbf{Academic Benefit}
& Q24\_7
& How frequently do you find that GenAI has helped improve your grades or performance in this course?
& Likert \\

& Q24\_8
& How frequently do you find that GenAI helps you retain course materials and achieve long-term learning goals?
& Likert \\

\textbf{Early Reliance}
& Q24\_9
& How frequently do you find that GenAI explains concepts better than your instructor?
& Likert \\

\textbf{GenAI Policy Clear}
& Q24\_10
& How frequently have you received clear guidelines about appropriate uses of GenAI in your classes?
& Likert \\

\textbf{Negative Impact}
& Q24\_11
& How frequently do you find that GenAI hurts your critical-thinking abilities?
& Likert \\

& Q24\_12
& How frequently do you feel scared to use GenAI?
& Likert \\

\textbf{Early Reliance}
& Q24\_13
& How frequently do you use GenAI before thinking about a problem by yourself?
& Likert \\

& Q24\_14
& How frequently do you use GenAI about a problem before reaching out to your instructor?
& Likert \\

\midrule

\textbf{Attitudinal Change}
& Q25\_1
& GenAI has changed how I interact with my instructors.
& Likert \\

& Q25\_2
& GenAI has changed how I interact with my peers.
& Likert \\

\textbf{GenAI No Impact}
& Q25\_3
& GenAI has had little impact on my overall learning experience.
& Likert \\

\textbf{Positive Impact}
& Q25\_4
& Overall, GenAI improved my learning in this class.
& Likert \\

\midrule

\textbf{Open Comments}
& Q26
& Do you have any comments for the researchers? If you liked something or disliked something, or something did not work for you, please mention it here.
& Open-ended \\

\end{longtable}

\normalsize

\end{document}